\documentclass{article} 
\usepackage{iclr2027_conference,times}

\usepackage{amsmath,amsfonts,bm}

\def\eqref#1{equation~\ref{#1}}

\def\1{\bm{1}}

\DeclareMathAlphabet{\mathsfit}{\encodingdefault}{\sfdefault}{m}{sl}
\SetMathAlphabet{\mathsfit}{bold}{\encodingdefault}{\sfdefault}{bx}{n}

\usepackage{hyperref}
\usepackage{url}
\usepackage{graphicx}
\usepackage{booktabs}
\usepackage{tcolorbox}

\title{SEDualVLN: A Spatially-Enhanced Dual-System Framework for Vision-Language Navigation}

\author{
\textbf{Jingzhi Huang}\textsuperscript{1} \hspace{0.9em}
\textbf{Junkai Huang}\textsuperscript{2} \hspace{0.9em}
\textbf{Wenxuan Song}\textsuperscript{3} \hspace{0.9em}
\textbf{Yuhang Peng}\textsuperscript{4} \hspace{0.9em}
\textbf{Hailong Huang}\textsuperscript{1} \and
\textbf{Haoyang Yang}\textsuperscript{1} \hspace{0.9em}
\textbf{Lujia Wang}\textsuperscript{3} \hspace{0.9em}
\textbf{Yi Wang}\textsuperscript{1}\textsuperscript{\dag} \\[1.5ex]
\textsuperscript{1}Hong Kong Polytechnic University \quad
\textsuperscript{2}Institute of Automation, Chinese Academy of Sciences \\
\textsuperscript{3}Hong Kong University of Science and Technology (Guangzhou) \quad
\textsuperscript{4}DeepRobotics \\[1ex]
\texttt{jingzhi.huang@connect.polyu.hk}, \texttt{yi-eie.wang@polyu.edu.hk} \\
\textsuperscript{\dag}Corresponding author \\
\url{https://github.com/kim-os/SEDualVLN}
}

\iclrfinalcopy 
\begin{document}

\maketitle
\fancyhead{} 
\renewcommand{\headrulewidth}{0pt} 
\fancyfoot[C]{\thepage} 

\begin{abstract}
Vision-Language Navigation (VLN) approaches have currently followed two primary paradigms: the end-to-end Vision-Language Model (VLM) policy fine-tuned on navigation trajectories to directly predict actions, and the zero-shot modular pipeline integrating pre-trained Multimodal Large Language Model (MLLM) for training-free generalization to unseen environments. However, end-to-end methods struggle with long-horizon navigation and lack dynamic reasoning ability, whereas zero-shot methods are constrained by limited spatial grounding for reliable planning and require substantial reasoning time. To bridge this gap, we introduce SEDualVLN, a spatially-enhanced dual-system VLN framework. System 1 is a VLM model enhanced with both global and local spatial awareness, tailored for high-frequency action generation. System 2 integrates a general MLLM with a mapping module, wherein the MLLM performs low-frequency waypoint planning by leveraging top-down views of the real-time 3D map alongside streams of rendered path images. Both systems exploit complementary spatial representations to improve the agent’s spatial understanding in VLN tasks. Ultimately, they cooperate to complete the navigation task through a fast-slow coordinated approach. SEDualVLN achieves state-of-the-art performance on VLN-CE benchmarks, and further ablation studies demonstrate the effectiveness of each system and module.
\end{abstract}

\section{Introduction}
\label{sec:1}

Vision-Language Navigation (VLN) \cite{gu-etal-2022-vision,zhang2024vln} is a challenging and crucial task in embodied AI, which requires agents to accept visual observation inputs and follow language instructions to navigate in an unseen environment. Recent advancements in Multimodal Large Language Model (MLLM) \cite{video-llava,zhang2410video,nwae403} and Vision-Language-Action (VLA) \cite{zhang2024uninav,cheng2024nav,Zheng_2024_CVPR} model have significantly increased the appeal of VLN. In particular, advancing explicit spatial reasoning to accurately align language instructions with continuous 3D geometry and environmental topology has emerged as a critical prerequisite for achieving robust, long-horizon navigation in complex settings.

Existing VLN methods are generally divided into two main paradigms: \textbf{the end-to-end paradigm} \cite{zhang2024uninav,cheng2024nav,zhang2024navid,goetting2024,Song_2025_CVPR,wei2025streamvln,zeng2025janusvln,liu2026dygeovln,li2026p3nav} and \textbf{the zero-shot paradigm} \cite{MSNav,yue2026spatialvln,Open-Nav,Constraint-Aware,spatialnav,SpatialGPT,himemvln,instructnav,dreamnav}. The end-to-end VLN paradigm typically employs a unified vision-language policy, fine-tuned on expert navigation trajectories to directly predict sequential actions. While effective in constrained or seen environments, it struggles with long-horizon adaptation and lacks the dynamic, step-wise reasoning required for complex multi-step decision-making. Conversely, zero-shot pipelines leverage pre-trained MLLM to enable training-free generalization across unseen scenes. However, their planning reliability is inherently constrained by the limited spatial grounding and geometric reasoning capabilities of off-the-shelf foundation models. Consequently, existing methodologies face complementary limitations that hinder robust navigation in complex open-world scenarios.

Although recent studies \cite{dualsystem,ZhongZZHGWLHGPH26} have proposed dual-system approaches to compensate for the limitations of individual-systems, these methods merely provide a coarse integration of the two fundamental paradigms, without fundamentally enhancing the capabilities of the underlying systems. Specifically, compared to semantic capabilities, spatial capabilities are the most abstract and easily overlooked in VLN tasks, yet spatial awareness can directly improve the agent’s robustness and success rate in long-horizon navigation.

First, to address the limitations of the single system in VLN, we introduce a dual-system with fast–slow cooperation to jointly complete navigation tasks. Next, to enhance the spatial awareness of the underlying systems in spatial navigation, we equip System 1 with Global and Local Spatial Enhancement Strategies, enabling the VLM model to comprehend the environmental structure from coarse to fine. In System 2, we adopt a three-stage ``Mapping–Rendering–Reasoning'' pipeline, which provides the MLLM with reasoning cues from both global and dynamic perspectives.

\begin{figure}[t] 
  \centering
  \includegraphics[width=0.80\textwidth]{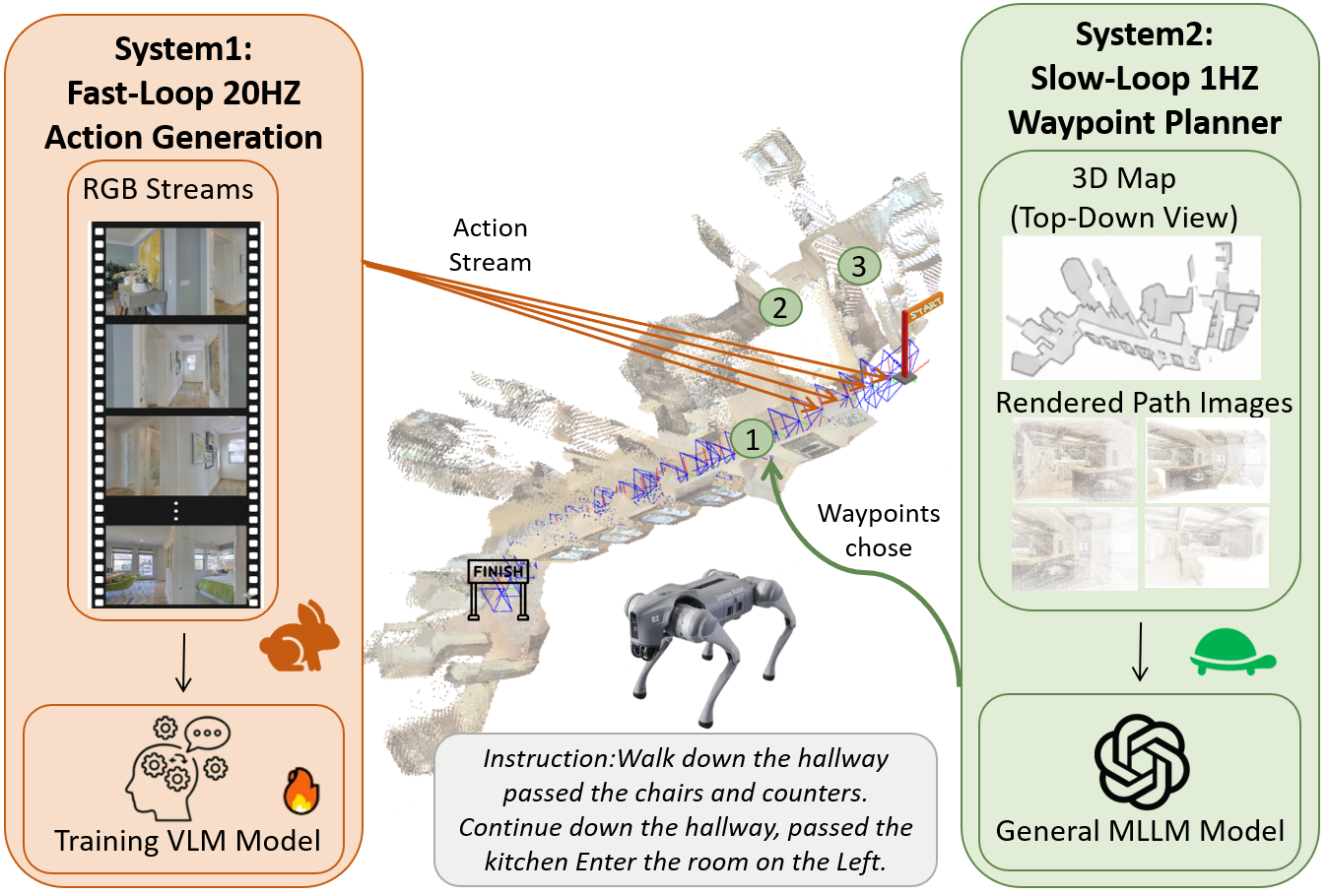} 
  \caption{Overview of SEDualVLN. System 1 (orange) generates fast, low-level actions from RGB streams via a VLM, while System 2 (green) plans waypoints using top-down 3D map and rendering path images through a general MLLM. The systems cooperate in a fast-slow coordinated loop.}
  \label{fig1} 
\end{figure}

Our contributions are as follows:
\begin{enumerate}
    \item \textbf{Dual-System VLN Framework:} We propose SEDualVLN, a dual-system architecture that combines a fast VLM-based action generator with a slow MLLM-based planner, coordinated by a scheduler, to jointly perform complex navigation tasks.
    
    \item \textbf{Spatially-Enhanced VLM (System 1):} We introduce Global and Local Spatial Enhancement strategies for the VLM model, enabling coarse-to-fine comprehension of the environment and improving spatial awareness for robust long-horizon navigation.
    
    \item \textbf{Mapping--Rendering--Reasoning MLLM (System 2):} We design a three-stage pipeline that provides the MLLM with both global and dynamic visual cues, enhancing spatial reasoning and waypoint planning in unseen environments.
    
    \item \textbf{Comprehensive Evaluation:} We demonstrate that SEDualVLN achieves state-of-the-art performance on VLN-CE benchmarks and conduct ablation studies to validate the effectiveness of each system and module in improving navigation success and robustness.
\end{enumerate}

\section{Related Work}
\label{related_work}
\subsection{Vision-Language Navigation (VLN)}
\label{sec:2.1}
Vision-Language Navigation (VLN) requires an embodied agent to follow natural language instructions to navigate in unseen photorealistic 3D environments \cite{Anderson,Qi_2020_CVPR,Krantz,rxr}. Early studies mainly focused on discrete settings (e.g.,R2R \cite{Anderson}), emphasizing cross-modal alignment via attention mechanisms and topological memory for long-term planning \cite{tan-etal,Hong_2021_CVPR}. Recent advances \cite{cheng2024nav,wei2025streamvln,dualsystem} have shifted toward continuous environments (e.g.,VLN-CE \cite{Krantz,rxr}), exploring pretraining, data augmentation, and online adaptation techniques to improve generalizability.

\textbf{End-to-end VLN approaches} integrate visual perception, language understanding, and action prediction in a unified framework by fine-tuning Video-LLMs \cite{zhang2410video,video-llava} or Vision-Language-Action (VLA) models on large-scale navigation data. These methods \cite{zhang2024navid,zhang2024uninav,cheng2024nav,wei2025streamvln} leverage video stream encoding, spatio-temporal context modeling, and techniques such as token pruning or DAgger \cite{DAgger} to generate low-level actions directly from RGB observations \cite{zhang2024navid,wei2025streamvln,liu2026dygeovln}. While achieving strong benchmark performance, they often struggle with long-horizon navigation due to accumulating context length and redundant historical token processing, and exhibit limited dynamic reasoning in complex scenes.

\textbf{Zero-shot VLN approaches} employ powerful multimodal LLMs (e.g., GPT-4o \cite{gpt4o}) as planners with carefully designed prompts and Chain-of-Thought reasoning, often in a training-free manner. Dual-process frameworks further combine lightweight expert models for routine navigation with LLMs for anomalous cases. These methods \cite{yue2026spatialvln,Constraint-Aware,ZhongZZHGWLHGPH26} excel at commonsense reasoning and generalization but lack fine-grained spatial and geometric understanding, leading to suboptimal decisions in precise spatial correlation and continuous video streams.

\subsection{Spatial Reasoning in VLN}
\label{sec:2.2}
Spatial reasoning is fundamental to VLN, requiring precise alignment between linguistic instructions, environmental topology, and continuous 3D geometry. Recent video-based VLMs and VLA models \cite{zhang2024navid,wei2025streamvln,zeng2025janusvln} enhance spatial retention through temporal context modeling and memory decoupling. However, their spatial representations remain largely implicit, making them susceptible to metric drift and poor interpretability in complex layouts. To mitigate this, explicit spatial reasoning has emerged as a critical direction. Topological planners \cite{Chen_2022_CVPR} construct evolving connectivity graphs with dual-scale transformers for global action planning, while geometry-aware methods \cite{Huo_2023_CVPR} integrate depth/normal maps with slot attention to learn geometry-enhanced visual representations. Dedicated spatial grounding frameworks like Spatial-VLN \cite{yue2026spatialvln} and SpatialNav \cite{spatialnav} explicitly model geometric constraints and spatial relations through structured perception and scene graphs. Concurrently, LLM-driven approaches \cite{ZhongZZHGWLHGPH26,dualsystem} decompose navigation into deliberative spatial planning via chain-of-thought or constraint-aware filtering. Despite their logical strength, these methods predominantly operate over abstract textual cues or fragmented 2D proxies, lacking direct access to continuous, physically consistent 3D geometry—frequently inducing spatial hallucination and degrading long-horizon robustness.

To bridge this gap, we propose a spatially-enhanced dual-system architecture. System 1 explicitly models spatial semantics and connectivity to establish a robust topological prior for efficient path grounding. System 2 leverages 3D mapping to render observed trajectories, providing high-fidelity, multi-view-consistent 3D context for deliberative planning. By unifying explicit topological reasoning with physically grounded 3D observation, our framework achieves more accurate, interpretable, and robust spatial reasoning compared to prior implicit or abstraction-dependent methods.

\section{Method}
\label{headings}

We formulate continuous VLN task as a sequential decision-making problem under monocular RGB perception. At each step $t$, the agent processes a language instruction $I$ alongside a trajectory of ego-centric observations $O_t = \{x_0, \dots, x_t\}$, where $x_\tau$ is the visual input at time $\tau$. To navigate successfully, the agent must learn a policy $\pi(O_t, I)$ to dictate the next atomic action $a_{t+1} \in \mathcal{A}$ based on its visual history and the given instruction. The discrete action space is restricted to $\mathcal{A} = \{\text{$Move Forward$}, \text{$Turn Left$}, \text{$Turn Right$}, \text{$Stop$}\}$. An episode is deemed a success if the agent issues a $Stop$ command within a proximity threshold of 3m from the target place, without exceeding a maximum budget of 500 steps.

As illustrated in Figure \ref{fig1}, SEDualVLN is a dual-system VLN framework. System 1 (Sec. \ref{sec:3.1}), is a VLM model that generates actions based on video stream input. System 2 (Sec. \ref{sec:3.2}), consists of a general MLLM model and mapping module that determines optimal navigation waypoints using 3D mapping overhead view and frontier-path rendering visualizations. Systems 1 and 2 operate at different frequencies to achieve a fast-slow coordinated approach: while System 2 requires more time for waypoint-level reasoning, System 1 executes multiple low-level actions during this period. Once System 2 completes its planning, the agent proceeds to execute the planned waypoints, thereby mitigating delays from lengthy reasoning while still incorporating global spatial information.

\begin{figure}[t] 
  \centering
  \includegraphics[width=0.80\columnwidth]{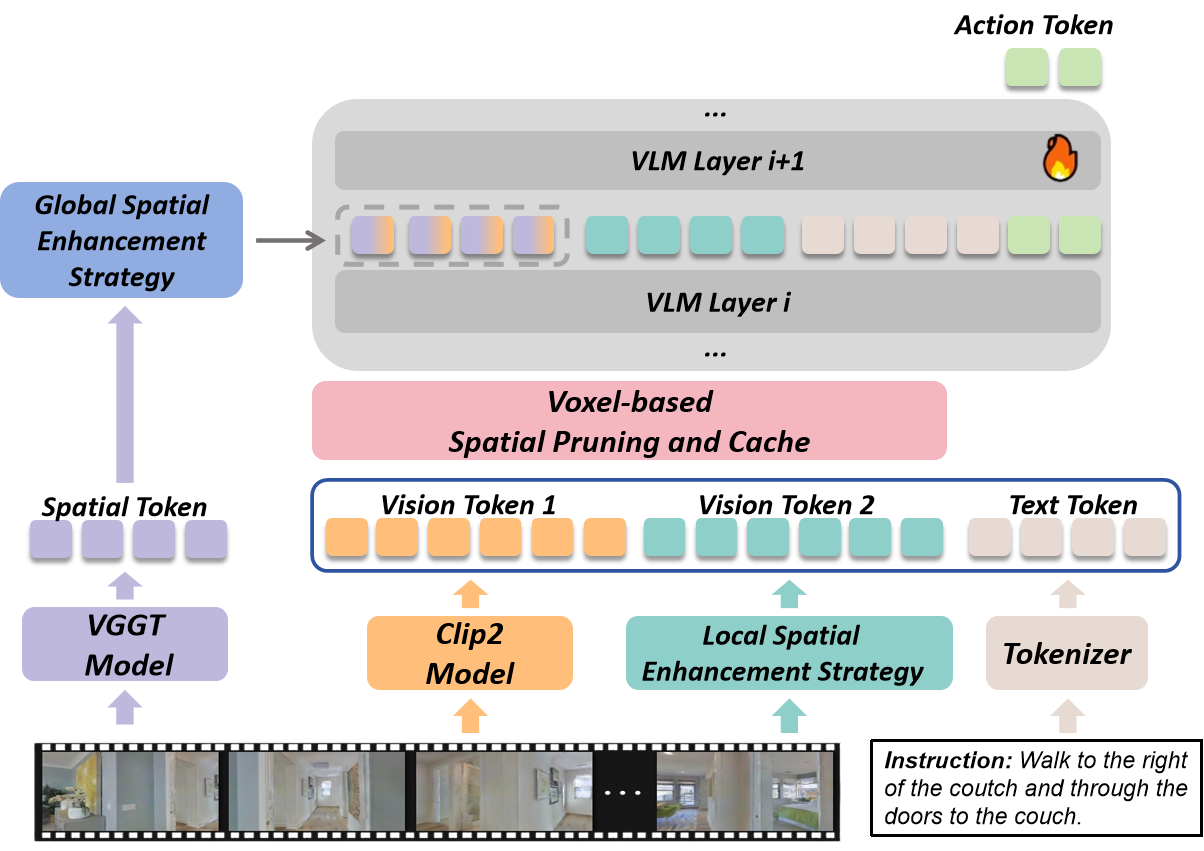} 
  \caption{Framework of System 1. Given RGB image streams and language instructions, the model extracts visual, spatial, and text tokens. During training, a frozen VGGT model provides feature-level spatial supervision for the visual tokens. The remaining tokens are processed by voxel-based spatial pruning and caching before being fed into the VLM.}
  \label{fig2} 
\end{figure}

\begin{figure}[t] 
  \centering
  \includegraphics[width=0.94\textwidth]{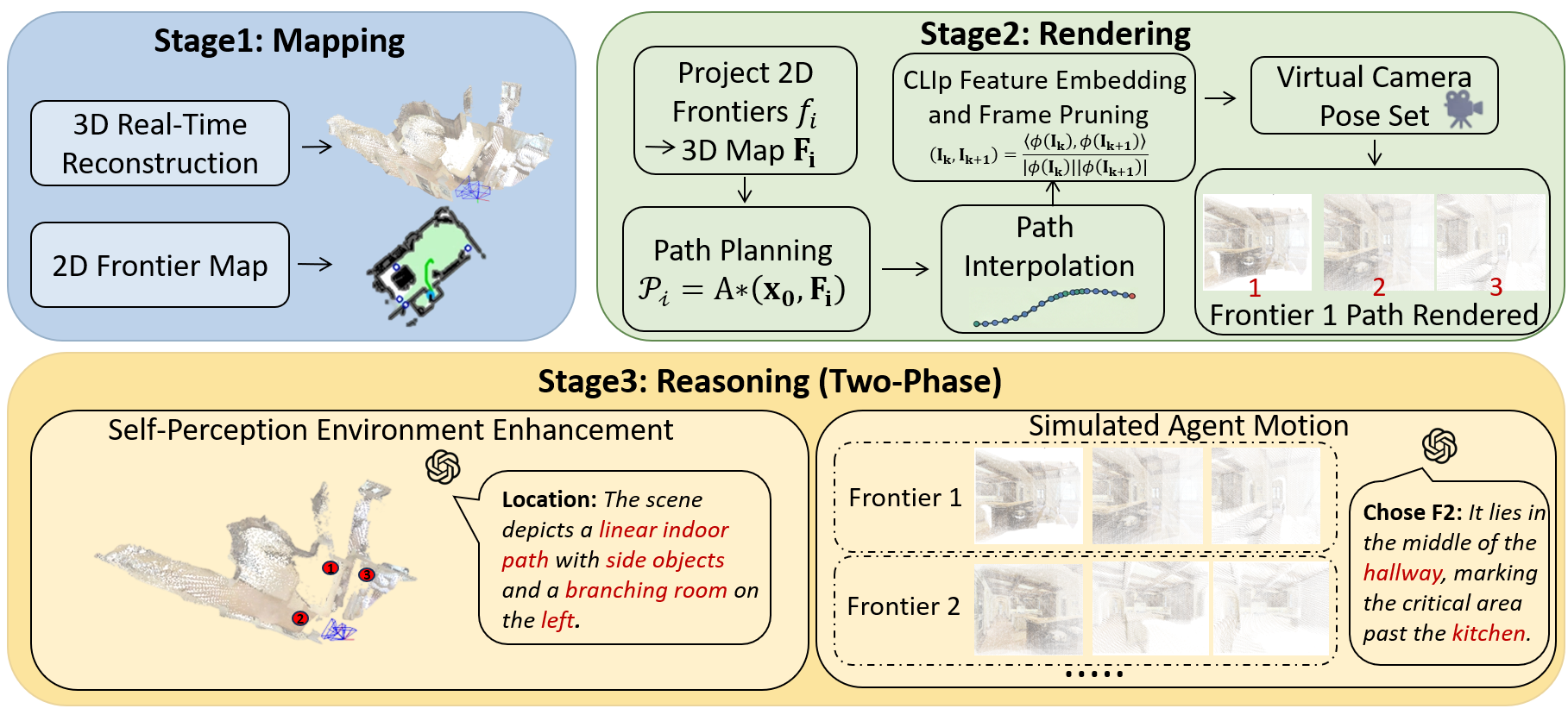} 
  \caption{System 2 Workflow. It consists of three stages: Mapping, which builds a 3D map and 2D frontier map; Rendering, which generates virtual camera views along planned paths; and Reasoning, where the agent first interprets the environment and then evaluates frontier paths against the instruction to select the optimal navigation target.}
  \label{fig3} 
\end{figure}

\subsection{System1: Spatially-Enhanced VLM Model}
\label{sec:3.1}
The overall framework of System 1 is illustrated in Figure \ref{fig2}. Our model is built upon the StreamVLN \cite{wei2025streamvln} backbone, adopting its multi-turn dialogue mechanism, KV cache, and spatial token pruning. We introduce two major modifications, enhancing its spatial awareness for long-range tasks.
First, we introduce a global spatial enhancement strategy that encourages the VLM’s visual tokens to encode spatial information without requiring explicit depth-map inputs. Second, we introduce Local Spatial Enhancement Strategy, enabling the VLM to focus more effectively on the connectivity of nearby passages and thereby reducing the risk of taking incorrect branches.

\paragraph{Global Spatial Enhancement Strategy}
Compared to approaches that provide explicit depth maps or reconstructed 3D geometry as inputs, we enhance spatial awareness through feature-level supervision. Specifically, during training, we use a frozen pretrained VGGT model to provide spatially aware feature targets for the VLM’s visual tokens. Prior work ~\cite{huang2024deciphering} has shown that supervising relatively deep but not the deepest layers is most effective for improving action performance, likely because the deeper layers lose more low-level visual features, making it more difficult for them to receive supervision targeted at spatial representations. Therefore, we apply supervision at intermediate attention fusion layers of LLaVA-Video ~\cite{zhang2410video} (the specific layers are selected based on ablation studies).

For each frame along the agent’s navigation sequence, visual token embeddings are projected through a two-layer MLP with batch normalization to align with the spatial representation dimension. During training, these projected embeddings are supervised against spatial features generated by a frozen pretrained VGGT model ~\cite{wang2025vggt}, with positional encodings added to preserve spatial ordering. The supervision is implemented by minimizing the cosine distance between the projected visual tokens and the corresponding spatial representations, encouraging the VLM to encode geometric and structural cues in its visual tokens. VGGT is used only to generate the training targets and is not required during inference. Formally, let $\mathbf{V}_t$ denote the projected visual token at frame $t$, $\mathbf{S}_t$ the spatial representation from the 3D model, and $\mathbf{p}_t$ the positional embedding; the combined training objective is defined as
\begin{equation}
\mathcal{L}_{\text{SE}} = \mathcal{L}_{\text{action}} + \alpha \cdot \frac{1}{N} \sum_{t=1}^{N} \Big[ 1 - \cos\big(\mathbf{V}_t, \mathbf{S}_t + \mathbf{p}_t\big) \Big],
\end{equation}
where $N$ is the number of frames and $\alpha$ balances the action prediction loss and the spatial supervision loss.

\paragraph{Local Spatial Enhancement Strategy.}
In VLN tasks, it is common for an agent to move from one region to another, where passages play a critical role as the connecting structures between regions. Most failures of VLM occur when the agent enters an incorrect passage, making it unable to return to the correct trajectory. Extracting global visual tokens from the current RGB frame using a 3D model is a common approach; however, such visual tokens often struggle to capture passage existence due to interference from numerous irrelevant tokens. To address this issue, we propose a channel connectivity extraction module, designed to enable the VLM to explicitly focus on passage connectivity. This module operates as follows: (1) the presence of passages (e.g., corridors, doors) is detected using an open-vocabulary Grounding DINO \cite{liu2024grounding} approach; (2) the detected passages are then segmented via SAM \cite{kirillov2023segment} to obtain binary masks; (3) all passage masks are set to 1 while non-passage regions are set to 0, completing the image encoding. Finally,  MLP is employed to project the resulting representations to the same dimensionality as the global visual tokens.

\subsection{System2: Mapping--Rendering--Reasoning MLLM}
\label{sec:3.2}

The overall workflow of System 2 is illustrated in Figure~\ref{fig3}, comprising three main stages: Mapping, Rendering, and Reasoning.

\begin{table}[t] 
\centering
\caption{Performance comparison of VLN methods on R2R and RxR Val-Unseen splits. * indicates methods using the waypoint predictor.}
\setlength{\tabcolsep}{0.55mm} 
\small 

\begin{tabular}{ccccccccccccc}
\toprule
Method & \multicolumn{4}{c}{Observation Encoder} & \multicolumn{4}{c}{R2R Val-Unseen} & \multicolumn{4}{c}{RxR Val-Unseen} \\
\cmidrule(lr){2-5} \cmidrule(lr){6-9} \cmidrule(lr){10-13}
 & Pano. & Odo. & Depth & S.RGB & NE$\downarrow$ & OS$\uparrow$ & SR$\uparrow$ & SPL$\uparrow$ & NE$\downarrow$ & SR$\uparrow$ & SPL$\uparrow$ & nDTW$\uparrow$ \\
\midrule
CMA* \cite{Hong_2022_CVPR} & \checkmark & \checkmark & \checkmark & & 6.20 & 52.0 & 41.0 & 36.0 & 8.76 & 26.5 & 22.1 & 47.0 \\
BERT* \cite{Hong_2022_CVPR} & \checkmark & \checkmark & \checkmark & & 5.74 & 53.0 & 44.0 & 39.0 & 8.98 & 27.0 & 22.6 & 46.7 \\
ETPNav* \cite{ETPNav} & \checkmark & \checkmark & \checkmark & & 4.71 & 65.0 & 57.0 & 49.0 & 5.64 & 54.7 & 44.8 & 61.9 \\
LAW\cite{LAW} & & \checkmark & \checkmark & \checkmark & 6.83 & 44.0 & 35.0 & 31.0 & 10.90 & 8.0 & 8.0 & 38.0 \\
Seq2Seq \cite{Krantz} & & & \checkmark  & \checkmark & 7.77 & 37.0 & 25.0 & 22.0 & 12.10 & 13.9 & 11.9 & 30.8 \\
\midrule
NaVILA \cite{cheng2024nav} & & & & \checkmark &  5.22 & 62.5 & 54.0 & 49.0 & 6.77 & 49.3 & 44.0 & 58.8 \\
StreamVLN \cite{wei2025streamvln} & & & & \checkmark  & 4.98 & 64.2 & 56.9 & 51.9 & 6.22 & 52.9 & 46.0 & 61.9 \\
DualVLN \cite{dualsystem} & & & & \checkmark  & 4.05 & 70.7 & 64.3 & 58.5 & 4.58 & 61.4 & 51.8 & 70.0 \\
\bfseries SEDualVLN(ours) & & & & \checkmark & \textbf{3.75} & \textbf{73.7} & \textbf{67.3} & \textbf{62.5} & \textbf{4.12} & \textbf{63.9} & \textbf{52.4} & \textbf{72.8} \\
\bottomrule
\end{tabular}
\label{Tab1}
\end{table}

\paragraph{Mapping.} We construct a lightweight 3D map of the environment and a 2D frontier map in real-time. The 2D frontier map guides the agent to explore new regions, following the construction rules detailed in VLFM ~\cite{yokoyama2024vlfm}. Multi-view 3D mapping is computationally expensive and requires many agent steps. To alleviate this, we adopt a feed-forward approach based on LingBot-Map ~\cite{chen2026geometric}, which incrementally builds the 3D map online.

\paragraph{Rendering.} Let $\mathcal{F} = \{\mathbf{f}_1, \dots, \mathbf{f}_n\}$ denote the set of frontier points in 2D coordinates. Each frontier point $\mathbf{f}_i$ is projected into the 3D map to obtain its corresponding 3D location $\mathbf{F}_i$. For each $\mathbf{F}_i$, we compute a collision-free path from the agent's current position $\mathbf{x}_0$ to $\mathbf{F}_i$ using the A* algorithm: $\mathcal{P}_i = \text{A*}(\mathbf{x}_0, \mathbf{F}_i)$.

To generate a dense set of rendering views along the path, we perform interpolation between consecutive path nodes. Specifically, if the Euclidean distance between two consecutive points $\mathbf{p}_j$ and $\mathbf{p}_{j+1}$ exceeds a threshold $d$, we insert intermediate points such that the spacing satisfies $\|\mathbf{p}_{j+1} - \mathbf{p}_j\| < d$. Each interpolated point corresponds to a virtual camera pose, and the RGB image at that pose is rendering as $\mathbf{I}_k$. To reduce redundant information and MLLM token usage, we apply a cosine-similarity based pruning:
\begin{equation}
s(\mathbf{I}_k, \mathbf{I}_{k+1}) = \frac{\langle \phi(\mathbf{I}_k), \phi(\mathbf{I}_{k+1}) \rangle}{\|\phi(\mathbf{I}_k)\|\|\phi(\mathbf{I}_{k+1})\|} < \tau \implies \text{keep } \mathbf{I}_{k+1},
\end{equation}
where $\phi(\cdot)$ denotes a CLIP \cite{radford2021learning} encoder embedding  and $\tau$ is a similarity threshold. After sampling and redundancy removal, we obtain $m$ RGB frames $\{\mathbf{I}_1, \dots, \mathbf{I}_m\}$ for each path.

\paragraph{Reasoning.} The reasoning stage operates in two phases. In the first phase, self-perception environment enhancement, the GPT-4o \cite{gpt4o} is provided with the 3D top-down map and is required to describe the agent's current location and nearby environment. In the second phase, simulated agent motion, the GPT-4o evaluates each frontier path by comparing the Rendering image sequences $\{\mathbf{I}_1, \dots, \mathbf{I}_m\}$ with the natural language instruction. It then selects the frontier point whose path best aligns with the instruction. This two-stage process allows the agent to integrate spatial context and visual guidance when making navigation decisions.

\section{Experiments}
\label{Experiment}

\subsection{Experimental Setup}
\label{sec:4.1}
\paragraph{Simulation Benchmark Setup.}
We evaluate our method on two general VLN-CE benchmarks, R2R-CE ~\cite{Krantz} and RxR-CE ~\cite{rxr}. Both datasets are based on Matterport3D reconstructions, with navigation episodes exceeding 10 meters. The agent is allowed to navigate freely using low-level actions, such as moving forward 0.25 meters or turning left 15 degrees, rather than relying on fixed navigation graph nodes. All experiments are evaluated using standard metrics, including Navigation Error (NE), Success Rate (SR), Oracle Success Rate (OS), and Success weighted by Path Length (SPL), normalized Dynamic Time Warping (nDTW), following prior works.

\paragraph{Base Models and Implementation Details.}
We build System 1 based on the LLaVA-Video ~\cite{zhang2410video} 7B model, which employs Qwen2-7B \cite{hui2024qwen2} as the language backbone. We perform several additional modifications: first, during training, we reconstruct the loss function by incorporating a spatial alignment loss; second, we use Local Spatial Enhancement Strategy and extract additional visual tokens from the RGB images. Initially, we fine-tune the model for one epoch using only the oracle VLN trajectories. Subsequently, we use this model to collect DAgger trajectories and continue training for another epoch using both the VLN data and general multimodal data. Each training step processes 128 video clips. The entire training procedure takes approximately 2000 A100 GPU hours. System 1 runs using a single RTX 4090 GPU. System 2 requires no training; it utilizes the pre-trained 3D model LingBot-Map \cite{chen2026geometric} for 3D mapping, and its deployment also occupies one RTX 4090 GPU.

\begin{table}[t] 
\centering
\caption{Effect of Global Spatial Enhancement Strategy on VLN Performance.}
\setlength{\tabcolsep}{1.7mm} 
\small 

\begin{tabular}{c c c c c c c c c c}
\toprule
Target Rep. & Layer & \multicolumn{4}{c}{R2R Val-Unseen} & \multicolumn{4}{c}{RxR Val-Unseen} \\
\cmidrule(lr){3-6} \cmidrule(lr){7-10}
 & & NE$\downarrow$ & OS$\uparrow$ & SR$\uparrow$ & SPL$\uparrow$ & NE$\downarrow$ & SR$\uparrow$ & SPL$\uparrow$ & nDTW$\uparrow$ \\
\midrule
$\times$ & $\times$ & 4.10 & 67.3 & 61.3 & 58.7 & 4.43 & 58.5 & 49.5 & 64.3 \\
VGGT & 1 & 3.95 & 72.5 & 66.4 & 60.4 & 4.23 & 62.8 & 49.1 & 70.5 \\
VGGT & 8 & 3.88 & 72.7 & 66.8 & 61.6 & 4.17 & 63.5 & 51.8 & 72.3 \\
VGGT & 16 & 3.80 & 73.4 & 67.0 & 62.3 & 4.15 & 63.7 & 52.1 & 72.5 \\
VGGT & \textbf{24} & \textbf{3.75} & \textbf{73.7} & \textbf{67.3} & \textbf{62.5} & \textbf{4.12} & \textbf{63.9} & \textbf{52.4} & \textbf{72.8} \\
VGGT & 32 & 3.84 & 72.8 & 66.7 & 61.7 & 4.18 & 63.4 & 51.9 & 72.6 \\
\bottomrule
\end{tabular}
\label{Tab2}
\end{table}

\begin{table}[t]
\centering
\caption{Effect of  Local Spatial Enhancement Strategy (LSES) on VLN Performance.}
{%
\begin{tabular}{c c c c c c c c c}
\toprule
Method & \multicolumn{4}{c}{R2R Val-Unseen} & \multicolumn{4}{c}{RxR Val-Unseen} \\
\cmidrule(lr){2-5} \cmidrule(lr){6-9}
 & NE$\downarrow$ & OS$\uparrow$ & SR$\uparrow$ & SPL$\uparrow$ & NE$\downarrow$ & SR$\uparrow$ & SPL$\uparrow$ & nDTW$\uparrow$ \\
\midrule
w/o LSES  & 3.95 & 70.3 & 63.8 & 61.5 & 4.23 & 59.8 & 51.0 & 66.5 \\
SEDualVLN (Full) 
 & \textbf{3.75} & \textbf{73.7} & \textbf{67.3} & \textbf{62.5} & \textbf{4.12} & \textbf{63.9} & \textbf{52.4} & \textbf{72.8} \\
\bottomrule
\end{tabular}%
}
\label{Tab3}
\end{table}

\subsection{Experiment Result}
\label{sec:4.2}

\paragraph{Quantitative analysis.}
As shown in Table\ref{Tab1}, we compare our SEDualVLN with previous state-of-the-art methods on the two splits of VLN-CE. These methods can be categorized into four types: (1) multi-sensor-based approaches (CMA \cite{Hong_2022_CVPR}, BERT \cite{Hong_2022_CVPR}, ETPNav \cite{ETPNav}, Seq2Seq \cite{Krantz}), (2) VLM methods using first-person RGB-D images (LAW \cite{LAW}), (3) VLM methods using only first-person RGB images (NaVILA \cite{cheng2024nav}, StreamVLN \cite{wei2025streamvln}), and (4) dual-system architecture methods (DualVLN \cite{dualsystem}). Experimental results show that, without relying on additional data sources, our method achieves a SR improvement of 3\% and 2.5\% over DualVLN on the two splits, respectively, reaching a new state-of-the-art level.

\paragraph{Qualitative analysis.}
\label{sec:4.2.2}
As shown in Figure \ref{fig4}, we compare  with the state-of-the-art method StreamVLN. The results demonstrate that our model significantly reduces the probability of taking wrong turns at junctions, indicating enhanced spatial awareness.
As shown in Figure \ref{fig5}, we visualize navigation and mapping visualization of our SEDualVLN. For more experimental visualization results, please refer to the supplementary video material.

\subsection{Ablation Study}
\label{sec:4.3}
To demonstrate that each system and module of SEDualVLN is indispensable, we conduct extensive ablation studies, including: (1) evaluating the effect of  spatial enhancement strategy at different attention layers of the VLM on the  navigation performance; (2) assessing the effect of System 1’s channel connectivity extraction module; (3) comparing the navigation results when System 1 and System 2 are run independently versus collaboratively; and (4) analyzing the impact of different general-purpose MLLM.

\begin{table}[ht]
\centering
\caption{Impact of System 1 and System 2 Collaboration on VLN.}
\setlength{\tabcolsep}{1mm} 
\small 

\begin{tabular}{c c c c c c c c c c c c} 
\toprule
Method & Freq. (S1:S2) & \multicolumn{5}{c}{R2R Val-Unseen} & \multicolumn{5}{c}{RxR Val-Unseen} \\
\cmidrule(lr){3-7} \cmidrule(lr){8-12}
 & & NE$\downarrow$ & OS$\uparrow$ & SR$\uparrow$ & SPL$\uparrow$ & AT$\downarrow$ & NE$\downarrow$ & SR$\uparrow$ & SPL$\uparrow$ & nDTW$\uparrow$ & AT$\downarrow$ \\
\midrule
(a) system1 & - & 4.25 & 68.2 & 62.8 & 58.7 & \textbf{48.9} & 4.43 & 58.5 & 49.5 & 64.3 & \textbf{53.6} \\
(b) system2 & - & 4.34 & 65.7 & 59.8 & 57.9 & 294.3 & 4.53 & 56.6 & 49.1 & 70.5 & 313.8 \\
(c) system1+system2 & 10:1 & 3.83 & 72.7 & 67.0 & 61.6 & 59.9 & 4.17 & 63.5 & 51.8 & 72.3 & 67.3 \\
(d) system1+system2 & \textbf{20:1} & \textbf{3.75} & \textbf{73.7} & \textbf{67.3} & \textbf{62.5} & 57.3 & \textbf{4.12} & \textbf{63.9} & \textbf{52.4} & \textbf{72.8} & 63.9 \\
(e) system1+system2 & 30:1 & 3.80 & 72.3 & 66.5 & 60.9 & 52.8 & 4.15 & 63.4 & 52.1 & 72.5 & 59.8 \\
\bottomrule
\end{tabular}
\label{Tab4}
\end{table}

\begin{table}[ht]
\centering
\caption{Effect of different general-purpose MLLM on VLN.}
\small 
\begin{tabular}{c c c c c c c c c}
\toprule
Method & \multicolumn{4}{c}{R2R Val-Unseen} & \multicolumn{4}{c}{RxR Val-Unseen} \\
\cmidrule(lr){2-5} \cmidrule(lr){6-9}
 & NE$\downarrow$ & OS$\uparrow$ & SR$\uparrow$ & SPL$\uparrow$ & NE$\downarrow$ & SR$\uparrow$ & SPL$\uparrow$ & nDTW$\uparrow$ \\
\midrule
Gemini \cite{gemini}  & 3.82 & 73.4 & 66.9 & 61.9 & 4.19 & 63.4 & 52.1 & 71.9 \\
Qwen2.5-VL \cite{qwen25vl}  & 3.78 & 73.6 & 67.1 & 62.1 & 4.17 & 63.7 & 52.2 & 72.4 \\
GPT-4o \cite{gpt4o} & \textbf{3.75} & \textbf{73.7} & \textbf{67.3} & \textbf{62.5} & \textbf{4.12} & \textbf{63.9} & \textbf{52.4} & \textbf{72.8} \\
\bottomrule
\end{tabular}
\label{Tab5}
\end{table}

\paragraph{Effect of  Global Spatial Enhancement Strategy at Different Attention Layers of the VLM.} We analyze whether the spatial enhancement strategy effectively improves the agent's global spatial awareness, and we further study its effect when supervised at different layers of the VLM model. The experiments are conducted on two datasets for both training and evaluation. As shown in Table~\ref{Tab2}, the results indicate that supervising the Global Spatial Enhancement Strategy at layer 24 yields the best performance. Specifically, compared to the model without supervision,  improving the SR by 6.0\% and SPL by 3.8\% on the R2R Val-Unseen split, and the SR by 14.4\% and SPL by 11.3\% on the RxR Val-Unseen split. These results indicate that supervising the Spatial Enhancement Strategy at mid-to-deep layers is most effective in enhancing the agent's spatial.

\paragraph{Effect of  Local Spatial Enhancement Strategy.}
As shown in Table~\ref{Tab3}. Incorporating Local Spatial Enhancement Strategy  (LSES) consistently improves navigation metrics across both R2R and RxR Val-Unseen sets. On R2R, NE is reduced by 0.20, SR improves by 3.5\%, SPL increases by 1.0\% , and OS rises by 3.4\%. For RxR, NE decreases by 0.11, SR improves by 4.1\%, and nDTW increases significantly by 6.3\%. The SPL  increases by 1.4\% Overall, these results suggest that agents benefit from LSES during navigation, and LSES significantly improves success rates at decision points such as intersections.

\paragraph{Impact of System 1 and System 2 Collaboration on VLN.}
As shown in Table~\ref{Tab4}, to further demonstrate the effectiveness of the dual-system approach, we introduce an additional metric, Average Time (AT, in seconds), which measures the mean time taken by the agent to complete a navigation eposide. Comparing (a) and (d), we observe that the dual-system achieves significantly higher success rates than the pure VLM system, with SR improvements of 4.5\% and 5.4\%, albeit at the cost of slightly increased time. Comparing (b) and (d), the dual-system outperforms the general MLLM single-system, with SR gains of 7.5\% and 7.3\%, while the average completion time is reduced by a factor of six. Additionally, we examine the metrics under different switching frequency ratios, as shown in (c), (d), and (e), and find that a frequency ratio of 20:1 allows our SEDualVLN to best balance SR and time efficiency.

\paragraph{Effect of different general-purpose MLLM on VLN.}
As shown in Table\ref{Tab5}, using Gemini results in two benchmarks SR of 66.9\% and 63.4\%. The Qwen2.5-VL model performs slightly better, achieving an SR of 67.1\% and  63.7\%. However, GPT-4o demonstrates the highest efficacy. This indicates that GPT-4o provides superior spatial reasoning and zero-shot generalization capabilities.

\begin{figure}[t] 
  \centering
  \includegraphics[width=0.94\textwidth]{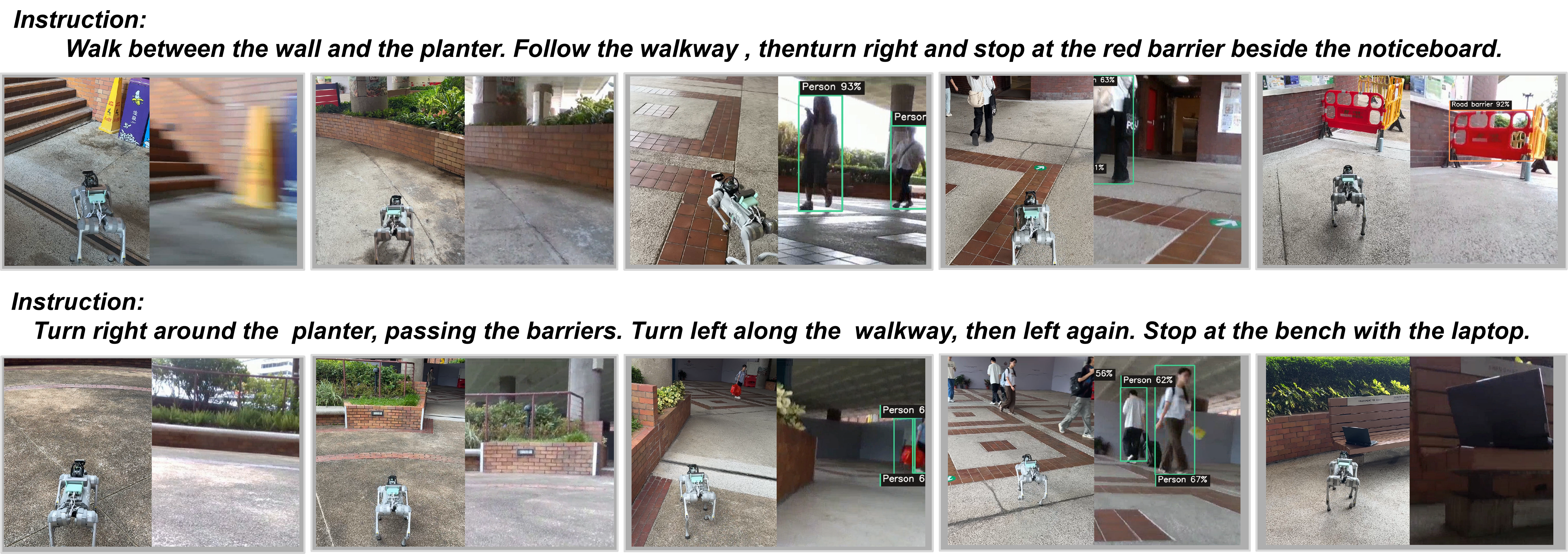} 
  \caption{Real-world experiments using the Unitree G2 robot. Guided by natural language instructions (top), SEDualVLN enables the robot to successfully navigate complex campus environments to reach specific targets (e.g., the red barrier and the bench). }
  \label{fig8} 
\end{figure}

\subsection{Real-World Experiments}
To validate the effectiveness of our proposed SEDualVLN, we deployed a Unitree G2 robot in complex campus environments with dynamic obstacles. As shown in Figure \ref{fig8}, driven by our SEDualVLN, the robot successfully executed multi-step natural language instructions (e.g., navigating to a red barrier or a bench) by accurately perceiving landmarks and avoiding pedestrians in real-time. By seamlessly integrating high-level semantic planning with low-level locomotion control, SEDualVLN enabled the robot to complete all tasks without human intervention. These real-world deployments conclusively demonstrate that our method effectively bridges the sim-to-real gap, ensuring robust and precise navigation in unstructured physical environments.

\section{Limitation}
\label{sec:5}
Our main limitation lies in deployment on physical devices. This is because the latest real-time 3D mapping features still face constraints in terms of real-world real-time performance, which leads to distorted 3D mapping results and consequently impacts the entire system.

\section{Conclusion}
\label{sec:6}
We presented SEDualVLN, a spatially-enhanced dual-system Vision-Language Navigation framework that successfully integrates the strengths of end-to-end VLM and zero-shot MLLM pipeline. To overcome the challenges of long-horizon navigation and limited spatial grounding, SEDualVLN employs a fast-slow coordinated approach. Both systems are explicitly designed to prioritize and enhance spatial awareness. Our results on the VLN-CE benchmarks establish a new state-of-the-art. By cultivating a stronger sense of direction and dynamic reasoning capabilities, SEDualVLN paves the way for more robust and capable navigation agents in complex, unseen environments.

\section*{AI use strategyment}
In this work, we used generative AI tools (e.g., large language models) for manuscript language editing, grammar refinement, and code debugging. We have not used generative AI tools for generating core research ideas, designing methodologies, or creating experimental data. We have thoroughly reviewed all AI-assisted text and code to ensure their accuracy, correctness, and originality. We take full responsibility for the final content of this work, including all text, claims, and artifacts.

\bibliography{iclr2027_conference}
\bibliographystyle{iclr2027_conference}

\newpage
\appendix
\section{Technical Details}
\label{Appendix}
In this section, we first build upon the previously mentioned qualitative experiment Section \ref{sec:4.2} to further illustrate how the two-stage reasoning of the MLLM in System 2 is performed. We then provide the prompts used in both System 1 and System 2. For complete video visualizations and additional experimental segments, please refer to the supplementary video material.

\begin{figure}[ht] 
  \centering
  \includegraphics[width=0.99\textwidth]{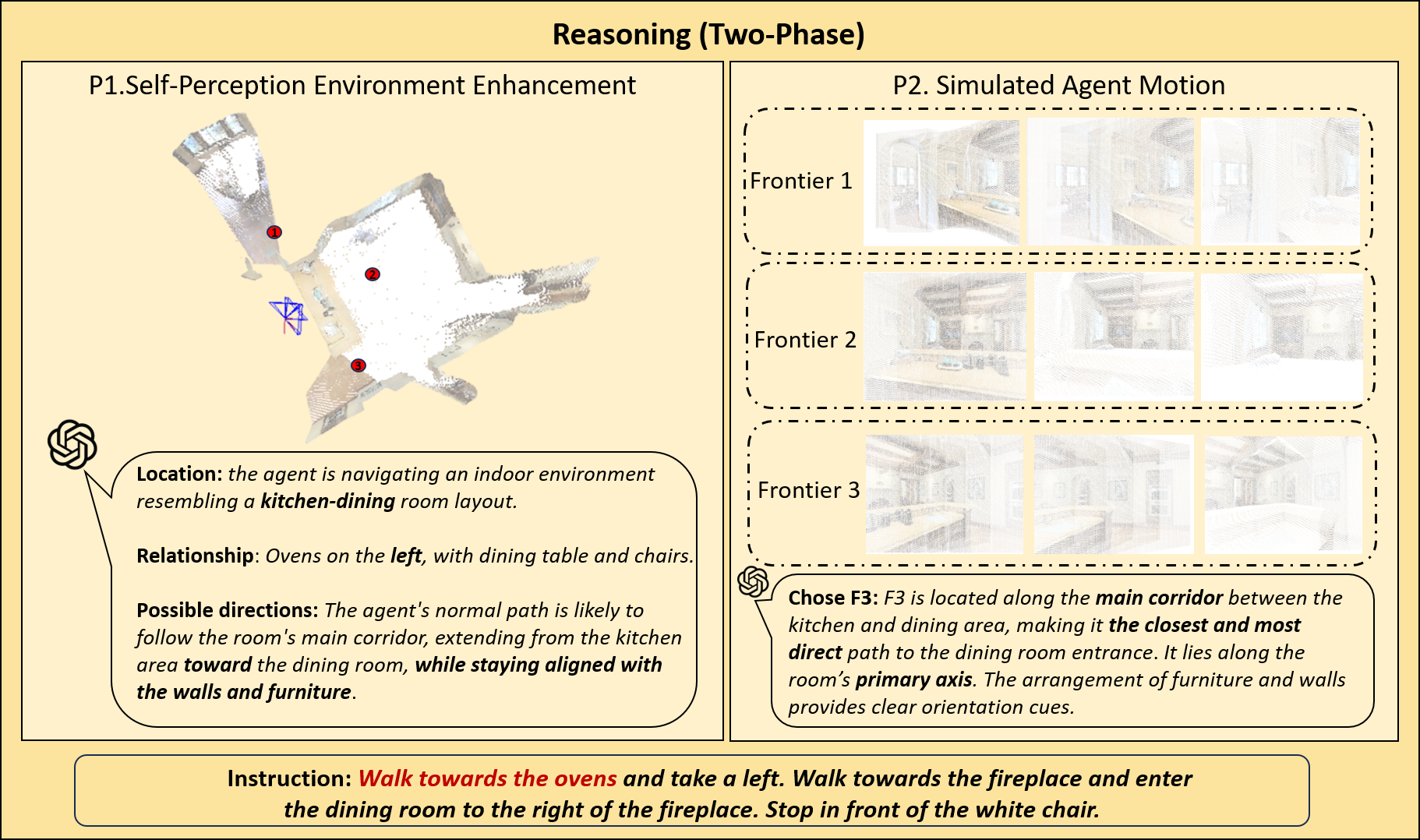} 
  \caption{The specific details of the MLLM reasoning are presented in the Case study.}
  \label{fig6} 
\end{figure}

\begin{figure}[ht] 
  \centering
  \includegraphics[width=0.99\textwidth]{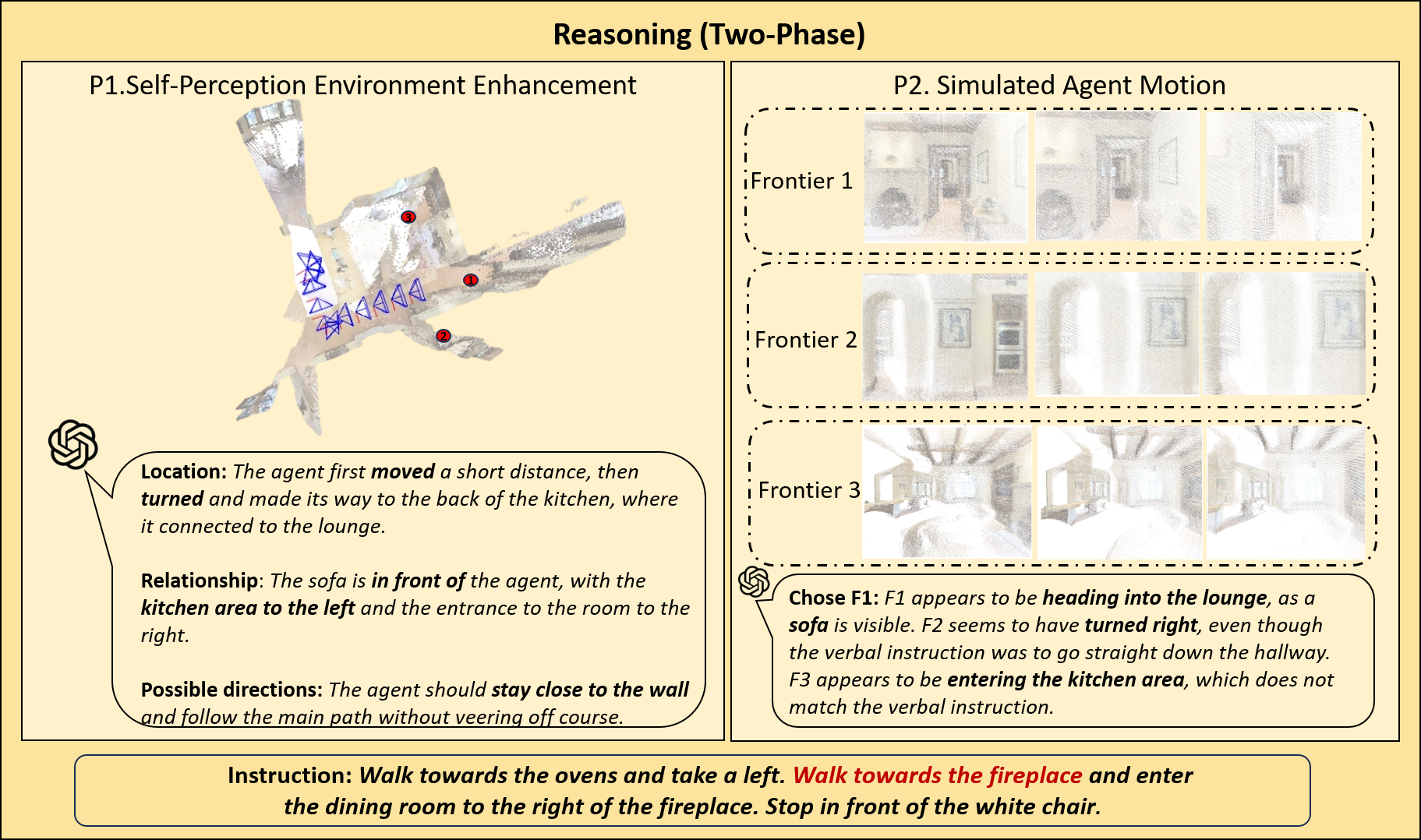} 
  \caption{The specific details of the MLLM reasoning are presented in the Case study.}
  \label{fig7} 
\end{figure}

As shown in Figure \ref{fig6} and Figure \ref{fig7}, the reasoning of the MLLM is divided into two stages. In the first stage, we guide the MLLM to analyze the environmental structure, object relationships, and possible orientations based on the top-down map. In the second stage, we provide the MLLM with additional cues through Rendering views, enabling it to complete the selection of waypoints.

Below are prompt templates for VLM and MLLM.

\begin{tcolorbox}[
    coltitle=white,         
    title=\large{\scriptsize VLM Prompts}, 
    arc=0mm,                
    boxrule=0mm,            
    toptitle=1mm,           
    bottomtitle=1mm,        
    left=4mm,               
    right=4mm,              
    top=2mm,                
    bottom=2mm,             
    fontupper=\itshape\scriptsize     
]  
You are an autonomous navigation assistant. Your task is to <Instruction>. Generate a sequence of actions to follow this instruction using the available commands: TURN LEFT (←) or TURN RIGHT (→) by 15 degrees, MOVE FORWARD (↑) by 25 centimeters, or STOP.

Input: [Images]\\

Model Output: JSON List \\
{[} \\
``Action'':  ``←↑↑→''\\
{]}
\end{tcolorbox}

\begin{tcolorbox}[
    coltitle=white,         
    title=\large{\scriptsize MLLM Reasoning Stage1 Prompts}, 
    arc=0mm,                
    boxrule=0mm,            
    toptitle=1mm,           
    bottomtitle=1mm,        
    left=4mm,               
    right=4mm,              
    top=2mm,                
    bottom=2mm,             
    fontupper=\itshape\scriptsize     
]  
    Task: You are an Environment Understanding Assistant. Analyze the 3D Top-Down map and the accompanying language instruction. Summarize the environment information for navigation planning.\\

Input: [3D Top-Down Map + Language Instruction]\\

Constraints:
\begin{itemize}
    \item Extract the key spatial information of the environment.
    \item Focus on \textbf{Location}, \textbf{Relationship}, and \textbf{Possible directions}.
    \item Keep the output concise and structured.
\end{itemize}

Model Output: JSON \\
{[} \\
``Location'': ``The agent is navigating an indoor environment resembling a kitchen-dining room layout.'', \\
``Relationship'': ``Ovens are located on the left side, with dining table and chairs positioned along the main corridor.'', \\
``Possible directions'': ``The agent's normal path is likely to follow the main corridor from the kitchen area toward the dining room, while staying aligned with walls and furniture.'' \\
{]}
\end{tcolorbox}

\begin{tcolorbox}[
    coltitle=white,         
    title=\large{\scriptsize MLLM Reasoning Stage2 Prompts}, 
    arc=0mm,                
    boxrule=0mm,            
    toptitle=1mm,           
    bottomtitle=1mm,        
    left=4mm,               
    right=4mm,              
    top=2mm,                
    bottom=2mm,             
    fontupper=\itshape\scriptsize     
]  
Task: You are a Navigation Planner Assistant. Analyze the images of each waypoint (Frontier 1, Frontier 2, Frontier 3, etc.) and select the most suitable waypoint for the agent to proceed. Provide clear reasoning for your choice based on visual cues such as corridor alignment, furniture arrangement, and spatial orientation.\\

Input: [Waypoint Images for Frontier 1, Frontier 2, Frontier 3, ...]\\

Constraints:
\begin{itemize}
    \item Evaluate each frontier marked in the images.
    \item Explicitly explain why the selected waypoint has the highest probability.
    \item Keep reasoning concise (2-3 sentences max).
    \item Do not suggest waypoints not shown in the images.
\end{itemize}

Model Output: JSON List \\
{[} \\
``Selected waypoint'': ``F3'', \\
``Reasoning'': ``F3 is along the main corridor between the kitchen and dining area, making it the closest and most direct path to the dining room entrance. It aligns with the room's primary axis, and the arrangement of furniture and walls provides clear orientation cues. Frontiers F1 and F2 lead to side areas with less clear access to the main corridor.'' \\
{]}
\end{tcolorbox}

\section{Experiment Details}
\label{Experiment Details}
\subsection{Qualitative Analysis}
\begin{figure}[h] 
  \centering
  \includegraphics[width=\columnwidth]{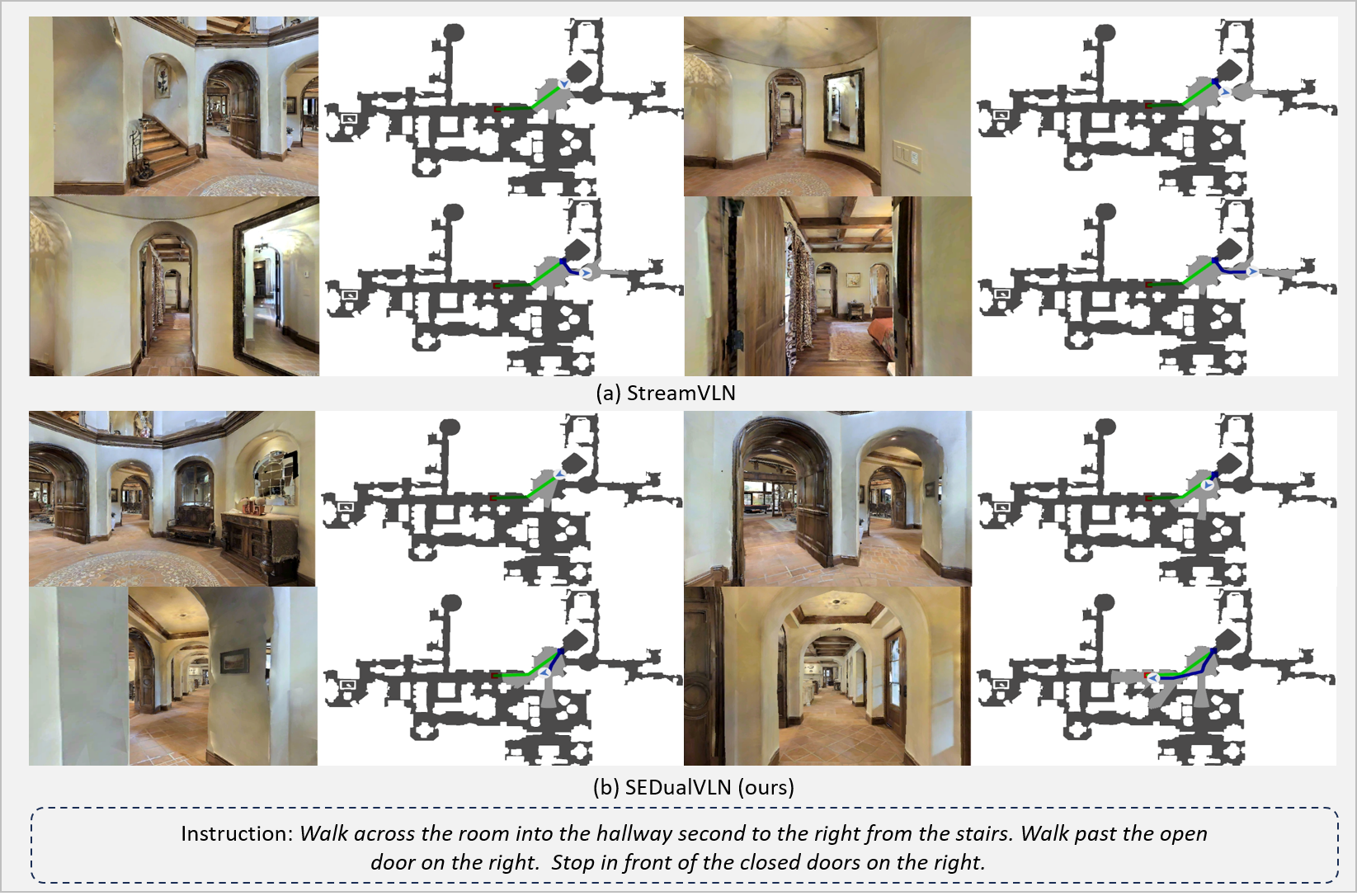} 
  \caption{Comparative experiment. SEDualVLN did not take a wrong turn at the beginning like StreamVLN; instead, it selected the correct path at the fork and successfully completed the task.}
  \label{fig4} 
\end{figure}

\begin{figure}[h] 
  \centering
  \includegraphics[width=\columnwidth]{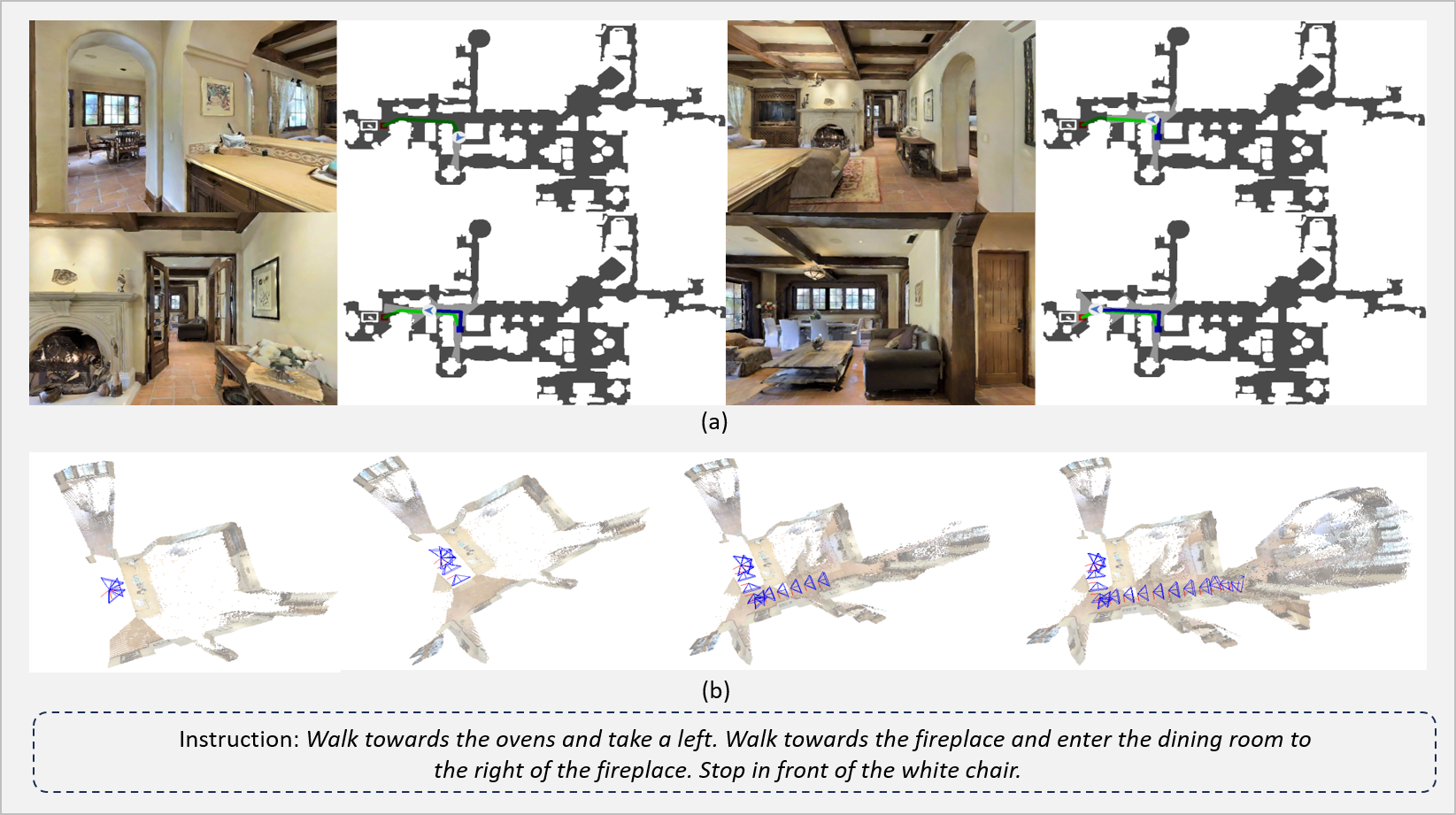} 
  \caption{Case study. Navigation and mapping visualization of our SEDualVLN.}
  \label{fig5} 
\end{figure}

\end{document}